\documentclass[letterpaper, 10 pt, conference]{ieeeconf}  % Comment this line out if you need a4paper

\IEEEoverridecommandlockouts                              % This command is only needed if 
\let\appendices\relax 
\usepackage{graphicx}
\usepackage{amsmath,amssymb}
\usepackage{cite}
\usepackage{subfigure}
\usepackage{algorithm}
\let\labelindent\relax
\usepackage{enumitem}
\newcommand{\subscript}[2]{$#1 _ #2$}
\usepackage{multirow}
\usepackage[title, page]{appendix}
\usepackage[noend]{algpseudocode}
\usepackage[font=footnotesize,labelfont=bf]{caption}

\title{\LARGE \bf
Seeking Visual Discomfort: Curiosity-driven Representations for Reinforcement Learning
}

\author{ Elie Aljalbout$^{1}$, Maximilian Ulmer$^{1}$ and Rudolph Triebel$^{1,2}$% <-this % stops a space
\thanks{$^{1}$Technical University of Munich (TUM), 80797 Munich, Germany 
        {\tt\small \{name.lastname@tum.de\}}} %
\thanks{$^{2}$Insitute of Robotics and Mechatronics, German Aerospace Center, 82234 Wessling, Germany
        {\tt\small Rudolph.Triebel@dlr.de}}%
}

\begin{document}

\maketitle
\thispagestyle{empty}
\pagestyle{empty}

%%%%%%%%%%%%%%%%%%%%%%%%%%%%%%%%%%%%%%%%%%%%%%%%%%%%%%%%%%%%%%%%%%%%%%%%%%%%%%%%
\begin{abstract}
Vision-based reinforcement learning (RL) is a promising approach to solve control tasks involving images as the main observation. State-of-the-art RL algorithms still struggle in terms of sample efficiency, especially when using image observations. This has led to increased attention on integrating state representation learning (SRL) techniques into the RL pipeline. Work in this field demonstrates a substantial improvement in sample efficiency among other benefits. However, to take full advantage of this paradigm, the quality of samples used for training plays a crucial role. More importantly, the diversity of these samples could affect the sample efficiency of vision-based RL, but also its generalization capability. In this work, we present an approach to improve sample diversity for state representation learning. Our method enhances the exploration capability of RL algorithms, by taking advantage of the SRL setup. Our experiments show that our proposed approach boosts the visitation of problematic states, improves the learned state representation, and outperforms the baselines for all tested environments. These results are most apparent for environments where the baseline methods struggle. Even in simple environments, our method stabilizes the training, reduces the reward variance, and promotes sample efficiency.
\end{abstract}

%%%%%%%%%%%%%%%%%%%%%%%%%%%%%%%%%%%%%%%%%%%%%%%%%%%%%%%%%%%%%%%%%%%%%%%%%%%%%%%%
\section{Introduction}
\label{sec:intro}
To solve complex tasks in unstructured environments, agents should be capable of learning new skills based on their understanding of their surroundings. Vision-based reinforcement learning is a promising technique to enable such an ability. These methods learn mappings from pixels to actions and may require millions of samples to converge, especially for physical control tasks~\cite{barth-maron2018distributional}. This sample inefficiency could be attributed to the complexity of the dynamics encountered in such environments, but also to the difficulty of processing raw image information. 

A recent paradigm to approach the latter problem is to enforce meaningful mid-level representations,  via integrating perception modules in the RL pipeline. These modules are trained either in a supervised~\cite{chen2020robust} or self-supervised/unsupervised fashion~\cite{de2018integrating}. While supervised methods are simpler and easier to train, they require access to labeled datasets, which are usually hard to obtain, especially for real-world robotics scenarios. Thus, unsupervised and self-supervised approaches are the most popular ones in recent work. In these settings, the main goal is to integrate state representation learning objectives in the RL process~\cite{ballard1987modular, yarats2019improving}. In contrast to end-to-end methods, approaches that leverage SRL explicitly encourage the policy to learn a state representation mapping based on observations. The additional objective improves sample efficiency as it provides an extra signal for training. However, during RL, the agent performs several trials to achieve a certain behavior. This trial and error process, together with the exploitative nature of RL algorithms could result in very similar samples being collected in the replay buffer. This lack of diversity can harm the generalization capability of the learned encoders and hinder the improvement in sample efficiency that could be achieved with SRL. Hence, data diversity could be very beneficial for vision-based RL, and exploration strategies tailored for diverse and SRL-problematic observations could boost sample efficiency even further. 

In this work, we aim at improving the sample diversity of vision-based RL. We present an approach for exploration that makes the agent specifically curious about the state representation. Our approach takes advantage of the off-policy property of most state-of-the-art RL algorithms and trains a separate curious policy based on the SRL error. A preliminary version of this work can be found in~\cite{Aljalbout2021making}. Our experiments show that the proposed method encourages the visitation of SRL-problematic states. Additionally, it improves the performance of downstream tasks, especially for environments where recent approaches struggle. It also stabilizes the training and reduces the reward variance for all environments. Our contributions can be summarized as follows:
\begin{itemize}[leftmargin=*]
    \item We present an approach for learning policies that are curious about the state representation.
    \item Our approach is independent of the choice of SRL methods.
    \item We demonstrate how the curious policy can be integrated into a vision-based RL pipeline.
    \item Our method improves the exploration, training stability, and overall performance of vision-based RL.
    \item Our approach enables learning previously unsolved vision-based RL tasks on the deepmind control suite (DMC).
    \item Our implementation will be made open-source upon publication.
    %\item We design our method with lifelong learning in mind.
\end{itemize}

%===============================================================================

\begin{figure*}
    \centering
    \includegraphics[width=0.9\linewidth]{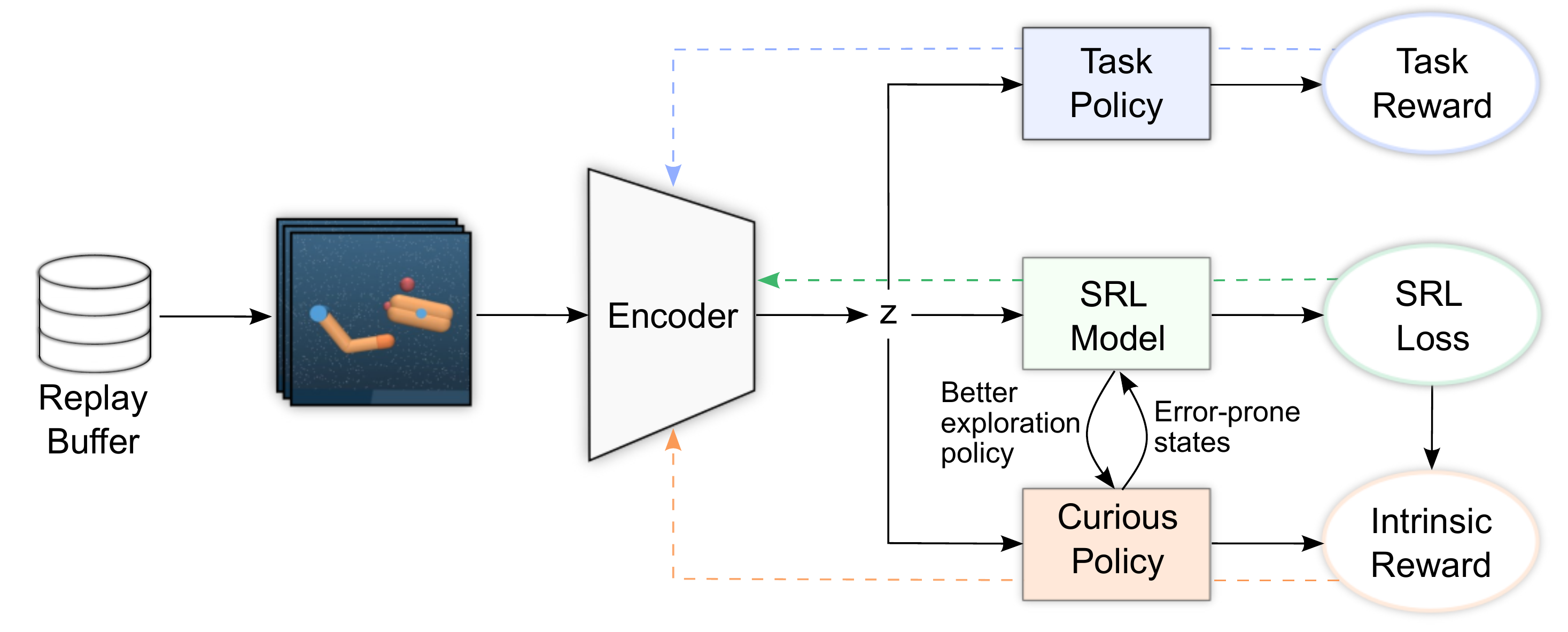}
    \caption{System Overview: our architecture is similar to the classical ones used for simultaneous state representation (SRL) and reinforcement learning (RL). Namely, an encoder is used to extract features from images and is trained together with an SRL model (e.g. decoder) to minimize the SRL loss. Simultaneously a task policy is trained to maximize the task reward, with the policy gradients flowing back to the encoder. In addition to the classical components, our method introduces a novel curious agent/policy, which is trained based on the SRL loss as an intrinsic reward. This creates an interplay between the SRL and the exhibited curious exploration behavior. The SRL guides the updates of the curiosity component, while the latter takes actions that lead to problematic and error-prone states. This in turn increases the diversity of observations.}
    \label{fig:Overview}
\end{figure*}

\section{Related Work}
\label{sec:rel_work}

\textbf{Integrating SRL.} One popular approach for SRL is the autoencoder (AE)~\cite{ballard1987modular}. One of the earliest works to integrate AEs in batch-RL can be found in~\cite{lange2010deep}. Later work explored the use of variational AEs~\cite{higgins2017darla} as well as regularized autoencoders (RAE)~\cite{yarats2019improving}. AEs could either be trained simultaneously with the policy~\cite{de2018integrating, yarats2019improving}, or in certain cases, separately pretrained before the RL start~\cite{higgins2017darla,lee2019making, aljalbout2020learning} or in an alternating fashion~\cite{lange2010deep}. More recent approaches take advantage of contrastive learning to boost the sample efficiency of vision-based RL. Namely, Laskin et al.~\cite{laskin2020curl} use data augmentations as positive samples and all other samples in the batch, as well as their augmentations as the negative ones. Similarly, the work in~\cite{stooke2020decoupling} uses contrastive learning to associate pairs of observations separated by a short time difference, hence uses (near) future observations as positive queries and all other samples in the batch as negative ones. Both AE-based methods and contrastive learning focus on compression of observation as the main goal for SRL. Another class of methods simultaneously learns feature encoders and dynamic models. For instance, Watter et al.~\cite{watter2015embed} impose locally linear transformations in the latent state space which enable long-term image predictions and control from raw images. Van Hoof et al.~\cite{van2016stable} present a very similar approach for RL based on both visual and tactile data. Besides compression of observations and latent dynamics models, Jonschkowski et al.~\cite{jonschkowski2015learning} presents an approach for SRL based on enforcing physical constraints such as proportionality, causality, repeatability, and temporal coherence. These constraints are formulated as objectives on the latent representations and are called robotic priors. 

In this work, we focus on compression-based approaches for SRL. Namely we use RAE and contrastive methods. Although contrastive learning has shown superior results to AE-based approaches~\cite{laskin2020curl, stooke2020decoupling}, these methods have many advantages. They are simple to implement, allow for integrating self-supervised objectives such as jigsaw puzzle~\cite{noroozi2016unsupervised}, enable multi-modal and multi-view fusion~\cite{lee2019making, akinola2020learning}, as well as task-specific objectives such as contact prediction~\cite{lee2019making}. More importantly, AE-based approaches lead to a more explainable state representation, especially when using generative AEs.
%%two vs one phase\\
%methods that use heuristics or demonstrations\\

\textbf{Curiosity in RL.} Classical deep RL algorithms work well in environments with rewards that are easy to encounter, but tend to fail once high-reward areas are harder to reach~\cite{burda2018exploration}.  This clearly motivates the use of exploration techniques as a means to achieve this goal. Popular paradigms for exploration include counts and pseudo-counts~\cite{bellemare2016unifying,ostrovski2017count}, learning distributions over value functions or policies~\cite{osband2016deep}, and information gain based methods~\cite{schmidhuber1991curious, houthooft2016vime,pathak2017curiosity}. While most of these approaches aim at improving the diversity in the replay buffer for improving the RL itself, our goal is to emphasize exploring states in which the SRL module struggles. This could help the agent in learning a more representative state space which would subsequently improve the RL. In addition, most previous approaches depend on the prediction error of a  dynamics model. In contrast, we leverage the SRL error for intrinsic motivation. This enables seamless integration of exploration in vision-based RL without any need for training additional dynamics models in the process. Furthermore, this creates an interplay between the SRL and the exploration which are both crucial aspects of successful and sample efficient vision-based RL. 

Most closely related to our method is the work in~\cite{seo2021state}. This work attempts to maximize state entropy using random convolutional encoders. The method uses a k-nearest neighbor entropy estimator in the representation space and uses this estimation as an additional intrinsic reward bonus for RL. Similar to our work, their approach doesn't require any dynamics models for training. However, using a k-nearest neighbor entropy estimator could be either compute-expensive if all observations need to be embedded at each step, or memory-expensive when those embeddings are saved in the replay buffer. Furthermore, a random encoder doesn't guarantee any notion of meaningful similarity between observations. In fact, in certain degenerate cases, the similarity in the representation space of a random encoder could be a measure of dissimilarity of the states. 
%abbeel paper with random encoders\\

\section{Background}
\textbf{Reinforcement Learning (RL)}  is a computational approach to automate policy learning by  maximizing  cumulative reward in an environment~\cite{sutton2018reinforcement}. RL tasks are usually formulated as Markov Decision Processes (MDP). A finite-horizon, discounted MDP is characterized by the tuple $\mathcal{M} = (\mathcal{S}, \mathcal{A}, \mathcal{P}, r, \rho_0, \gamma, T )$, where the state and action spaces are respectively $\mathcal{S}$ and $\mathcal{A}$ , transition dynamics $\mathcal{P}: \mathcal{S} \times \mathcal{A} \to \mathcal{S}$, reward $r: \mathcal{S} \times \mathcal{A} \to \mathbb{R} $, an initial state distribution $\rho_0$, discount factor $\gamma \in [0,1]$, and horizon $T$. The optimal policy $\pi : \mathcal{S} \to P(\mathcal{A})$, maximizes the expected discounted reward:
\begin{equation}
\label{eq:RL}
J(\pi) =  E_{\pi} \left[ \sum_{t=0}^{T-1} \gamma^t r(s_t, a_t) \right]
\end{equation}

\textbf{State Representation Learning (SRL).} While representation learning methods focus on learning abstract features from observations, SRL aims at learning low-dimensional features as state representations that are suitable for control. Namely, the goal of SRL is to learn a mapping from observations to state representations $g: \mathcal{O} \to \mathcal{Z}$, where $\mathcal{O}$ is the observation space and $\mathcal{Z}$ the embedding space. The mapping can also have as input a history of observations~\cite{lesort2018state}. In fully observable environments, SRL methods could attempt to recover the true state (depending on its definition). However, in partially observable setups, SRL aims at finding latent representations of the state. 

In this work, we focus on two methods for SRL. The first one is the regularized autoencoder (RAE)~\cite{Ghosh2020From}. RAE was introduced as a deterministic alternative to variational autoencoders (VAE)~\cite{Kingma2014,higgins2016beta}. It is trained using the following loss:

\begin{equation}
\label{eq:rae}
    \mathcal{L}_{SRL}(RAE) = \mathbb{E}_{o\sim D}[\log p_\theta(o|z) + \lambda_z ||z||^2 + \lambda_\theta||\theta||^2 ] 
\end{equation}

RAEs preserve the regularization properties of VAEs by explicitly penalizing the learned representation $z=g_\phi(o)$ and the decoder weights $\theta$,
where $o$ is the image observation, $g_\phi$ is the encoder, and $\lambda_z$, $\lambda_\theta$ are hyperparameters which respectively specify the influence of the $L_2$ penalty on $z$ and the weights decay for the decoder parameters. We choose RAEs over other autoencoder methods as they were previously shown to yield better performance when integrated with vision-based RL~\cite{yarats2019improving}. 

The second method we consider is based on contrastive learning (CL)~\cite{oord2018representation,pmlr-v119-henaff20a}. CL approaches learn representations based on similarity constraints pushing similar (positive) samples to be closer in the representation space and dissimilar (negative) ones to be further apart. In this work, we use the InfoNCE loss for CL~\cite{oord2018representation}:
\begin{equation}
\label{eq:CL}
    \mathcal{L}_{SRL}(CL) = \log \frac{\exp{q^TWk_+}}{\exp{q^TWk_+} + \sum_{i=0}^{K-1}q^TWk_i}
\end{equation}
where $q$ is the anchor, $\{k_i\}^K$ are all the targets including one positive $k_+$ and $K-1$ negatives. We follow the work in~\cite{laskin2020curl}, and use instance discrimination~\cite{wu2018unsupervised} for generating positive and negative keys. This means that the anchor and positive samples are augmentations of the same observation, while the negatives correspond to all other samples in the batch. Similar to~\cite{laskin2020curl}, we use random crops as the main source of augmentations. 
\section{ Learning Curiosity-Driven Representations}
\subsection{General Formulation}
Recent work on vision-based RL leverage SRL objectives to improve the sample efficiency of policy search methods~\cite{de2018integrating,yarats2019improving, laskin2020curl}. There are two main ways to integrate SRL in reinforcement learning. The first one is to simultaneously update both objectives, and the second is to train the two modules in an alternating fashion~\cite{de2018integrating}. The second option could also mean that SRL is only used to train a feature encoder in a pretraining phase preceding the actual RL. In both cases. the quality of the learned encoder and the resulting representations play a central role in the downstream RL tasks. With a finite amount of data, it is not always possible to collect enough samples to learn a representation that is valid across the state subspace relevant to the task at hand. For instance, in environments with sparse rewards, the SRL training rarely encounters observations corresponding to high-reward regions and their surroundings. The resulting representations for such observations might lack the necessary information for the policy to learn any useful behavior. 

In general, this lack of coverage is mostly attributed to the exploitative nature of RL algorithms, which leads to the replay buffer containing a lot of redundant and similar observations. Hence, to improve the quality of the feature extraction and learned representations, it is important to encourage collecting data in states outside of the comfort zone of the SRL model. Formally, that would correspond to maximizing the expected SRL error in the replay buffer $D$:
\begin{equation}
\label{eq:srlm}
 \max E_{p_o}  [\mathcal{L}_{SRL}(o)]
\end{equation}
where $p_o$ is the distribution of the observations in the replay buffer and $o$ is an observation.

\subsection{CuRe: Curiosity-Driven Representations}
Observations in the replay buffer are part of observation-action trajectories of length $T$ with  $p(o_0,a_0,...o_T,a_T)=p(o_0)\prod_{t=0}^T\pi(a_t|o_t)p(o_{t+1}|o_t,a_t)$ (we omit the generative process $p(o_t|s_t)$ for simplicity). Hence, only the initial state distribution and policy are relevant for (\ref{eq:srlm}), since the system dynamics are dependent on the environment and cannot be altered. In RL settings, the initial state distribution is dependent on the environment resetting mechanism. Although it's interesting to study the effect of this mechanism on performance, we leave this for future work. 

Instead, in this paper, we learn curious policies that maximize (\ref{eq:srlm}). This corresponds to training a policy $\pi_{cure}$ to maximize the objective in (\ref{eq:RL}) with $\gamma=0$ and the weighted SRL error as an intrinsic reward $r_{cure}=\beta \mathcal{L}_{SRL}$. $\beta$ is a hyperparameter that specifies the degree of curiosity. In our experiments we fix $\beta=1$. Furthermore, we allow $\gamma$ to have values different than zero as it didn't show any negative influence on the training.

\subsection{Integrating CuRe in RL}
There are two main ways to integrate CuRe in a vision-based RL algorithm. Namely, the intrinsic reward could either be added to the task reward $r_{task}$ to train the main policy $\pi_{task}$, or used separately for training a separate curious policy $\pi_{cure}$. Previous methods mostly use the earlier approach to integrate intrinsic rewards (based on a dynamics model)~\cite{pathak2017curiosity}. 

In this work, we choose the option with two separate policies. By doing so, we ensure that the task policy is purely optimizing the task reward, and additionally obtain a representation-curious agent capable of exploration for similar tasks in the same environment. More importantly, this choice allows our method to be used with both simultaneous and alternating approaches to SRL integration in RL. In addition, our early experiments indicate that a separate curious policy leads to substantially higher reward areas, while the single policy approach could deteriorate the results in comparison to the baselines. Furthermore, adding the rewards together usually introduces extra hyperparameters to weigh the different terms (e.g. in~\cite{pairet2019learning}, 3 extra hyperparameters are needed). It is important to note that having a separate policy is only possible when using off-policy RL algorithms such as soft-actor-critic (SAC)~\cite{haarnoja2018soft}, which is why we use this method in this work.

Our overall approach is illustrated in Figure~\ref{fig:Overview}. CuRe is agnostic to the choice of the SRL algorithm. Besides the encoder and the two policies, our architecture includes an SRL model. This model could refer to different modules depending on the SRL approach used. For instance, when using an AE-based method, it would correspond to a decoder. It could also refer to a dynamics model, an identity transformation, as well as any computational block that is used by representation learning methods to constrain the latent space. Furthermore, the updates of both policies affect the encoder parameters $\phi$. The SRL model parameters $\theta$  are only affected by the SRL update. 
At every step, we either sample actions from the main policy or the curious one. The choice of which policy to use at every step is based on a hyperparameter $p_{c}$ which specifies the percentage of times exploration actions should be sampled.  Intuitively, the curious policy is trained to reach states which have high SRL error. By occasionally sampling actions from this policy, the replay buffer ends up containing more problematic and diverse samples which helps to learn a better representation and to avoid overfitting. This interaction between the curious policy and the SRL model/loss results in an interplay similar to the one observed in generative adversarial networks~\cite{goodfellow2014generative}, as both modules are mutually beneficial to each other, and are trained in an adversarial setting. This interplay is illustrated in figure~\ref{fig:Overview}. The overall approach is summarized in algorithm~\ref{pseudo}. 

\begin{algorithm}
\caption{}
%\caption{
%CuRe: \textbf{Cu}riosity-Driven \textbf{Re}presentations for RL} 
\label{pseudo}
\begin{algorithmic}
\For{ each timestep $t=1...T$}
    \State $\epsilon \sim U(0,1)$ %\Comment{sample random number}
    \If{$\epsilon < p_c$} %\Comment{when below the exploration hyperparameter}
        \State $a_t \sim \pi_{cure}(.|o_t)$ %\Comment{sample action from curious policy}
    \Else %\Comment{otherwise}
        \State $a_t \sim \pi_{task}(.|o_t)$ %\Comment{sample action from task policy}
    \EndIf
    \State $o_{t+1}\sim p(.|o_t,a_t)$ %\Comment{step the environment}
    \State $D \gets D\cup (o_t,a_t,r_{task}(o_t,a_t),o_{t+1})$ %\Comment{add new tuple to the replay buffer}
    \State $B \gets SampleBatch(D)$ %\Comment{sample batch from replay buffer}
    \State $r_{cure} \gets UpdateSRL(B)$ %\Comment{using equation~(\ref{eq:rae}) or~(\ref{eq:CL})}
    \State $UpdateTaskAC(B)$ %\Comment{train task policy based on $r_{task}$}
    \State $UpdateCuriousAC(B, r_{cure})$ %\Comment{based on  $r_{cure}$}
\EndFor
\end{algorithmic}
\end{algorithm}

\section{Experiments} %reacher_easy, cartpole_swingup, ball_in_cup_catch, finger_spin, finger_turn, reacher_hard

We design experiments to answer the following questions:

\begin{enumerate}[label=(\subscript{Q}{{\arabic*}}), leftmargin=*]
%\setlength{\itemsep}{0.5pt}%
%\begin{itemize}
    \item Can we train a curious policy to increase the visitation of high SRL error states? 
    \item How does CuRe affect the performance, sample efficiency, and training stability of vision-based RL methods? Can CuRe be successfully integrated with multiple SRL methods?
    \item Does CuRe-driven SRL pretraining improve the performance of vision-based RL on downstream tasks?
%\end{itemize}
\end{enumerate}

\begin{figure}
    \centering
    \includegraphics[width=\linewidth]{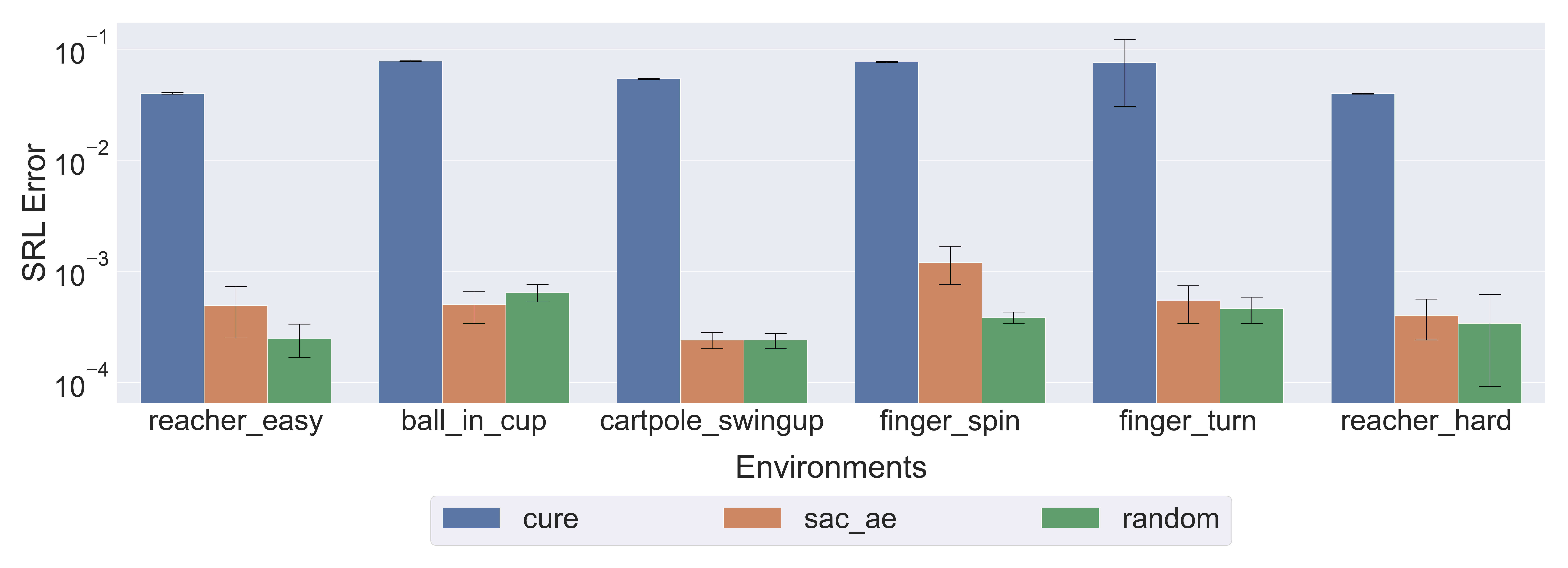}
    \caption{State representation learning (SRL) error encountered in trajectories sampled with three different policies: \textit{random}, \textit{sac\_ae} and our curious policy (\textit{cure}). The bars represent the mean error per step. The error bars represent the minimum and maximum encountered errors. Our method leads to the visitation of high SRL error states, around two orders of magnitudes more than the random and task policies (\textit{sac\_ae}).}
    \label{fig:ae_error}
\end{figure}

\begin{figure*}
    \centering
    %\subfigure[Finger Spin]{\includegraphics[width=0.32\textwidth]{figures/finger_spin.png}}
    %\subfigure[Cartpole Swingup]{\includegraphics[width=0.32\textwidth]{figures/cartpole_swingup.png}}
    %\subfigure[Ball in Cup Catch]{\includegraphics[width=0.32\textwidth]{figures/ball_in_cup_catch.png}}
    %\subfigure[Reacher Easy]{\includegraphics[width=0.32\textwidth]{figures/reacher_easy.png}}
    %\subfigure[Reacher Hard]{\includegraphics[width=0.32\textwidth]{figures/reacher_hard_new.png}}
    %\subfigure[Finger Turn]{\includegraphics[width=0.32\textwidth]{figures/finger_turn_new.png}}
    \includegraphics[width=0.9\linewidth]{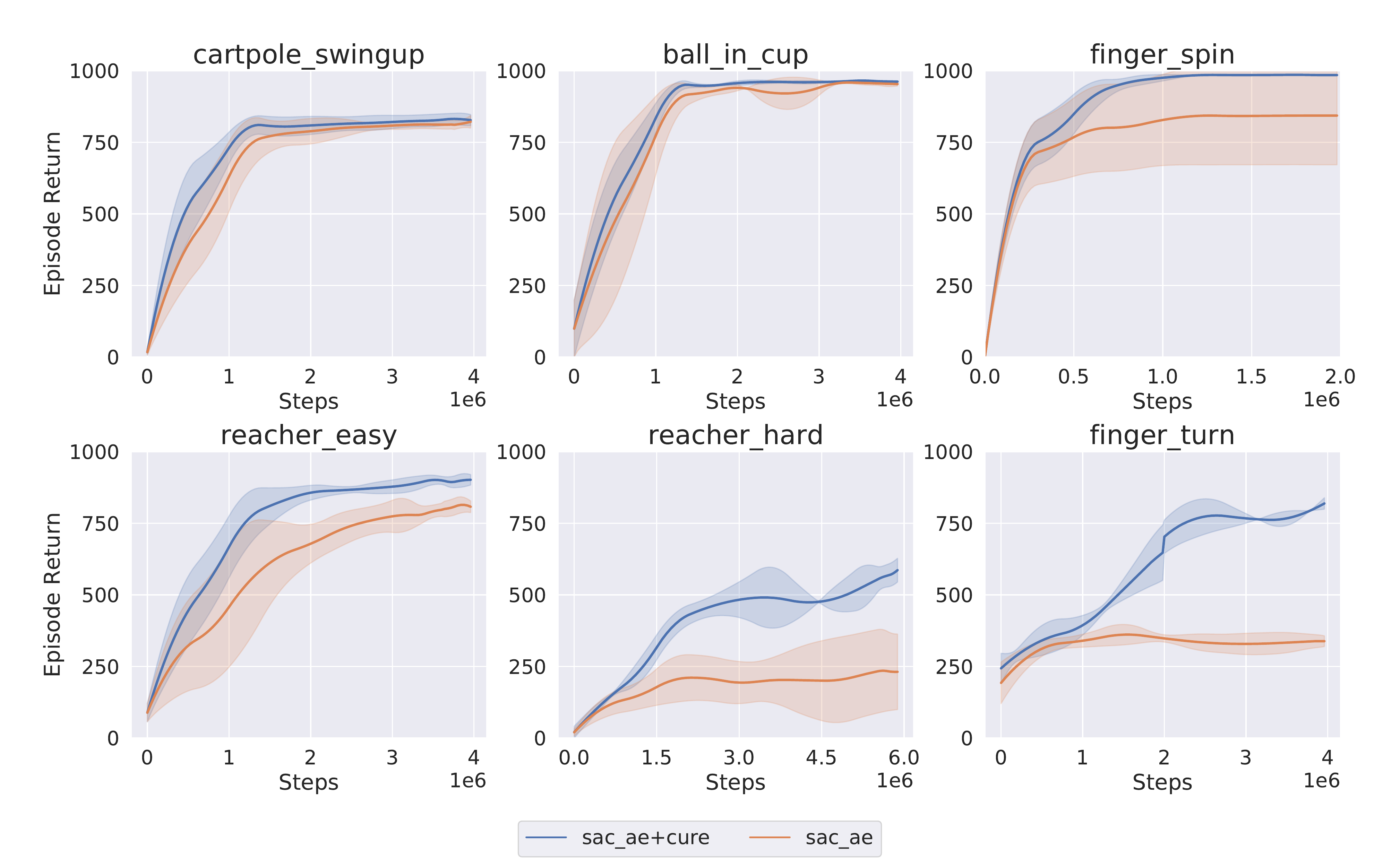}
    \caption{Training curves on six continuous control tasks from the Deepmind Control Suite~\cite{tassa2018deepmind}. The plots show the mean episode rewards of two algorithms. The first one is a baseline (\textit{sac\_ae}). The second method combines the same baseline with CuRe (\textit{sac\_ae+cure}). In all environments, our method exceeds the performance of the baseline. For easier tasks, the curious exploration either stabilizes the training or improves the maximum achieved reward. For the more difficult tasks, such as \texttt{finger\textunderscore spin}, \texttt{finger\textunderscore turn} and \texttt{reacher\textunderscore hard}, the additional curiosity objective allows to improve the average reward, where the baseline fails to reach high-reward areas.}
    \label{fig:results}
\end{figure*}

\subsection{Setup \& Baselines} 

To answer these questions, we experimentally evaluate our method on six continuous control tasks from the Deepmind Control Suite~\cite{tassa2018deepmind}. The chosen tasks aim to cover a wide range of common RL challenges, such as contact dynamics and sparse rewards. The tasks we use are \texttt{reacher\_easy}, \texttt{cartpole\_swingup}, \texttt{ball\_in\_cup}, \texttt{finger\_spin}, \texttt{finger\_turn} and \texttt{reacher\_hard}. As deep learning models could be energy inefficient~\cite{strubell2019energy}, we use only subsets of these tasks for minor experiments that are only aimed at validating simple aspects of our method.  

The main goal of our experiments is to validate the effectiveness of CuRe on improving the performance of already existing SRL-based approaches to vision-based RL. To do so we use two such algorithms as baselines and compare their performance with and without CuRe.
To validate, that the method is agnostic to the choice of SRL algorithms, we experiment with two different methods. Namely we use a combination of SAC with  RAEs as in \textit{sac\_ae}~\cite{yarats2019improving} and a combination of SAC with contrastive learning based on \textit{curl}~\cite{laskin2020curl}. %That way, we can validate the underlying hypothesis that CuRe is capable of augmenting SRL methods in RL to improve their SRL-specific exploration and subsequently the overall performance. 
We chose those two SRL methods since their integration in RL is fairly recent while also being well-established in robotics applications. We refrain from comparing our approach to classical exploration methods, since the two have different goals: classical exploration in RL is concerned with improving the sample diversity for RL while our method is aimed at encouraging the visitation of SRL-problematic states (discomfort zones). Hence comparing methods from these two categories could be misleading.
Both baselines and our method are implemented using PyTorch~\cite{NEURIPS2019_9015}. %Our implementation is based on the original \textit{sac\_ae} implementation~\cite{yarats2019improving}. 
For simplicity, we use the same hyperparameters for all experiments except for the action repeat value which changes per task, according to~\cite{hafner2019learning}. The actor and critic networks for the RL agent and the curious agent are trained using the Adam optimizer~\cite{kingma2014adam}, using default parameters. We store trajectories of experiences in a standard replay buffer. For implementing SAC, we follow the training procedure detailed in~\cite{yarats2019improving}. For the sake of reproducibility, we provide more information about the training procedure, and an overview of the hyperparameters in Appendix~\ref{app:hyper}. 
Our experiments required a training period of over four months on 6 GPUs (NVIDIA RTX2080 and RTX3090).

\subsection{Results}
\label{sec:results}

%\paragraph{Answering (\subscript{\textbf{Q}}{1})~:}
\textbf{Visiting High SRL Error Regions.}
Figure~\ref{fig:ae_error} shows the SRL error encountered when sampling actions from three different policies. The first policy generates \textit{random} actions within the action space of the environment. The second one is trained with \textit{sac\_ae}, and the last one is a CuRe-based curious policy that maximizes the SRL error without a task reward. While \textit{random} and \textit{sac\_ae} have similar mean errors per step, our method leads to the visitation of states which have on average an SRL error that is around two orders of magnitude higher. This confirms that CuRe fulfills its goal of increasing the probability of visiting high SRL error states. The exact values are shown in Table~\ref{table:ae_error} in the appendix. %In the following experiments, we'll look at the effect of this property on RL performance. 

\begin{figure*}
    \centering
    \includegraphics[width=0.915\linewidth]{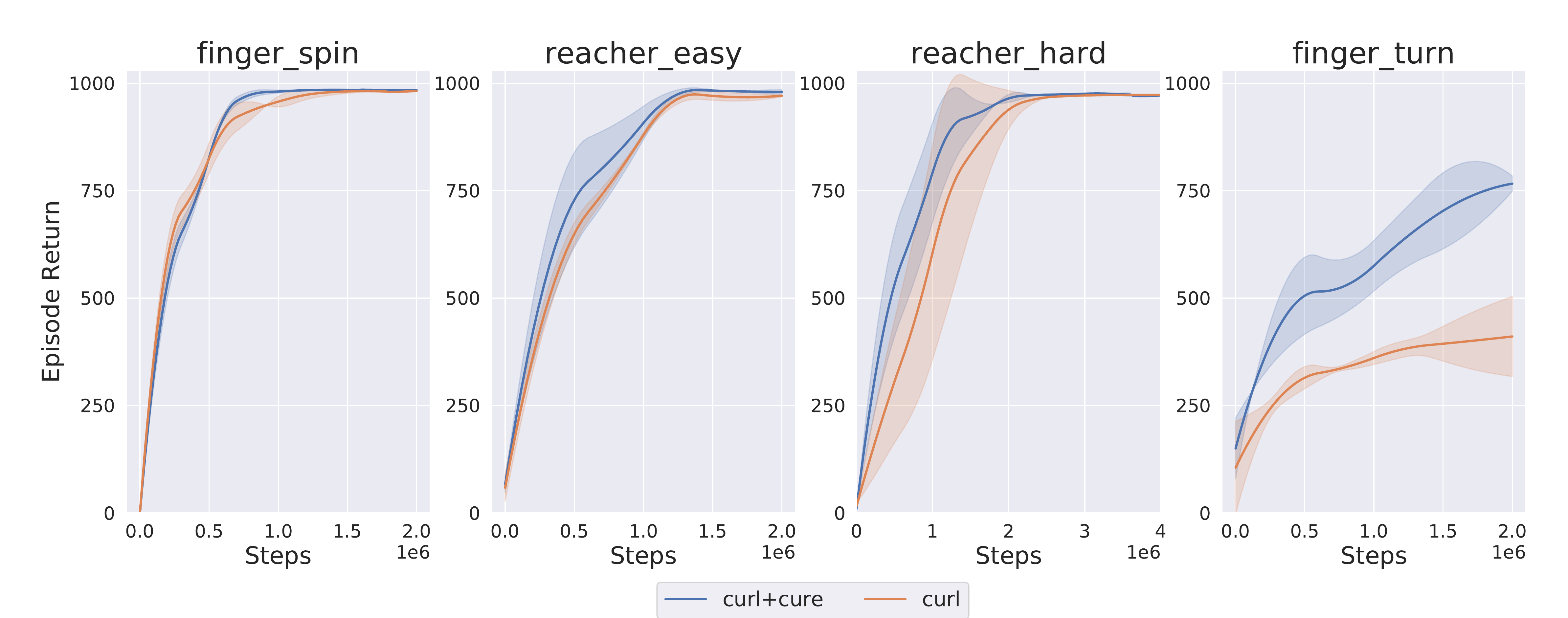}
    \caption{Training curves on four continuous control tasks from the Deepmind Control Suite~\cite{tassa2018deepmind}. The plots show the episode rewards of two algorithms. The first one is a baseline (\textit{curl}). The second method combines the same baseline with our curious policy (\textit{curl+cure}) . In all environments, our method improves the overall performance, sample efficiency and reward variance.}
    \label{fig:results__curl}
\end{figure*}

\begin{table*}[!htb]
  \centering
  \caption{Comparison of the performance (in terms of episode reward) of different versions of \textit{sac\_ae}: vanilla is the original algorithm~\cite{yarats2019improving}, random-pretraining and CuRe-pretraining refer to the cases where the vanilla procedure is preceded by an RAE pretraining phase using data collected with a random policy and a CuRe-based policy respectively.} \label{tab:pretraining}
  \resizebox{\textwidth}{!}{
  \begin{tabular}{lcccccc}
  \hline
\textbf{methods} & \textbf{cartpole\_swingup} & \textbf{ball\_in\_cup} & \textbf{finger\_spin} & \textbf{reacher\_easy} & \textbf{reacher\_hard} & \textbf{finger\_turn} \\ \hline
 \rule{0pt}{2.5ex}vanilla & 833$\pm$ 27 &953 $\pm$ 4   &820 $\pm$ 144   &714 $\pm$ 113  &169 $\pm$ 179  &229 $\pm$ 135   \\
 random-pretraining &784 $\pm$ 12   &\textbf{955} $\pm$ \textbf{10}   &975 $\pm$ 3 & 615 $\pm$ 129   &84 $\pm$ 33   &256 $\pm$ 40  \\ 
 CuRe-pretraining &\textbf{846} $\pm$ \textbf{25}  &504 $\pm$ 187   &\textbf{981} $\pm$ \textbf{7}  &\textbf{804} $\pm$ \textbf{52}   &\textbf{431} $\pm$ \textbf{40}   &\textbf{402} $\pm$ \textbf{58}   \\ 
 \hline
 \\

\end{tabular}
}
\end{table*}

%\paragraph{Answering (\subscript{\textbf{Q}}{3})~:}
\textbf{CuRe-based Exploration During RL.}
To answer (\subscript{Q}{2}), we study the effect of integrating CuRe into two different baselines, namely \textit{sac\_ae} and \textit{curl}. The integration is based on algorithm~\ref{pseudo}.
Figure~\ref{fig:results} shows the task reward for \textit{sac\_ae} with and without CuRe. In all environments, our method exceeds the performance of the baseline. Specifically, for tasks where the baseline doesn't show any signs of improvement, such as \texttt{reacher\textunderscore hard} and \texttt{finger\textunderscore turn}, CuRe leads to exploring high-reward areas, as can be seen when looking at the maximum rewards achieved in those environments. For simpler tasks such as \texttt{reacher\textunderscore easy} and \texttt{finger\textunderscore spin}, our method approaches the maximum environment rewards, while \textit{sac\_ae} converges to $80\%$. In addition, CuRe stabilizes the training and reduces the reward variance significantly. This last feature is not given enough attention in RL research. However, in real-world scenarios, when deploying RL agents, there could be cases where only one training run is possible. An algorithm with lower reward variance could guarantee a sufficiently good policy, while it's hard to say the same when this condition fails. This effect can also be seen for \texttt{cartpole\textunderscore swingup} and \texttt{ball\textunderscore in\textunderscore cup}. We observe that CuRe has a minor effect on the maximum reached reward for these last two environments. This could be attributed to the already good performance of the baseline on these tasks. In fact, in these environments, \textit{sac\_ae} already approaches the performance achieved by SAC trained with the true states~\cite{yarats2019improving}. Nonetheless, the additional curious exploration objective accelerates the convergence of all evaluation tasks, thus improving the sample efficiency, which is one key limitation of state-of-the-art model-free algorithms. In general, our experiments show that CuRe becomes more effective when the task complexity increases. %To validate the extent of this statement, additional experiments on more complex tasks are needed.

To study the effect of CuRe on \textit{curl}~\cite{laskin2020curl}, we run experiments on the four environments where CuRe had the most influence on \textit{sac\_ae}. Figure~\ref{fig:results__curl} shows the reward plots for \textit{curl} with and without CuRe. Similar to our previous results, CuRe has a positive impact on the overall performance, sample efficiency, reward variance and stability of training. This improvement is not as big as the one observed in our \textit{sac\_ae} experiments. However, this difference is understandable, since \textit{curl} is a more recent algorithm and has previously shown better results on similar deepmind control suite tasks~\cite{laskin2020curl}. Despite that, when looking at results on \texttt{finger\textunderscore turn} (Figures~\ref{fig:results} and ~\ref{fig:results__curl}), CuRe applied to \textit{sac\_ae} reaches a higher final episode reward than vanilla \textit{curl}. Additionally, we observe that \textit{sac\_ae+cure} has a better sample efficiency than \textit{curl} in the \texttt{finger\textunderscore spin} environment.

%\paragraph{Answering (\subscript{\textbf{Q}}{2})~:}
\textbf{Effect of Pretraining.}
In addition to our main results, to assess the quality of the learned representation with CuRe, and to answer (\subscript{Q}{3}), we study the effect of two different pretraining procedures on \textit{sac\_ae}. Namely, we look at pretraining the RAE using samples collected either using a random policy (random-pretraining) or using a policy trained with CuRe only, without any task reward (CuRe-pretraining). For both options, we perform the pretraining for $500$ thousand steps. We also compare the performance of those two variants to the case where no pretraining is performed at all~(vanilla). The results are shown in Table~\ref{tab:pretraining}. For all six environments, the best results are obtained when using one of the two pretraining mechanisms. In most cases, CuRe-based pretraining leads to better performance than random-pretraining. This become especially apparent for tasks where the vanilla method struggles, such as \texttt{reacher\textunderscore hard} and \texttt{finger\textunderscore turn}. However, for the \texttt{ball\textunderscore in\textunderscore cup} environment, CuRe-pretraining seems to deteriorate the performance when compared to both vanilla and random-pretraining. This could be attributed to the simplicity of the task, which reduces the need for SRL and SRL-tailored exploration. %Further ablation studies are needed to better understand this behavior. 
In general, although CuRe is beneficial for both SRL pretraining (Table~\ref{tab:pretraining}) and RL (figure~\ref{fig:results}), we observe that it is more effective during task learning than in the pretraining phase.

\section{Conclusion}
We introduce CuRe, a curiosity-based exploration technique that can be easily used together with state representation learning methods used in RL. This method exploits the SRL error to incentivize visiting more diverse and problematic states. We extensively evaluate our method on complex continuous control tasks in simulation. Our results show that our curious exploration method improves the performance of vision-based RL based on two different SRL methods. When comparing the baselines to the curiosity-driven extensions, we show that the added curiosity improves the performance in terms of speed of convergence, stability, and the total achieved reward. In future work, we plan to experiment with the transfer learning capability of our architecture, scale it up to multi-modal tasks and perform real-world experiments. %Furthermore, we plan to apply our method to real-world multimodal manipulation tasks.

%===============================================================================

% The maximum paper length is 8 pages excluding references and acknowledgements, and 10 pages including references and acknowledgements

% The acknowledgments are automatically included only in the final and preprint versions of the paper.

%===============================================================================

%\clearpage
\bibliographystyle{plain}
\bibliography{bibliography}  % .bib

\begin{thebibliography}{10}

\bibitem{akinola2020learning}
Iretiayo Akinola, Jacob Varley, and Dmitry Kalashnikov.
\newblock Learning precise 3d manipulation from multiple uncalibrated cameras.
\newblock In {\em 2020 IEEE International Conference on Robotics and Automation
  (ICRA)}, pages 4616--4622. IEEE, 2020.

\bibitem{aljalbout2020learning}
Elie Aljalbout, Ji~Chen, Konstantin Ritt, Maximilian Ulmer, and Sami Haddadin.
\newblock Learning vision-based reactive policies for obstacle avoidance.
\newblock {\em arXiv preprint arXiv:2010.16298}, 2020.

\bibitem{Aljalbout2021making}
Elie Aljalbout, Maximilian Ulmer, and Rudolph Triebel.
\newblock Making curiosity explicit in vision-based rl.
\newblock {\em arXiv preprint arXiv:2109.13588}, 2021.

\bibitem{ballard1987modular}
Dana~H Ballard.
\newblock Modular learning in neural networks.
\newblock In {\em AAAI}, pages 279--284, 1987.

\bibitem{barth-maron2018distributional}
Gabriel Barth-Maron, Matthew~W. Hoffman, David Budden, Will Dabney, Dan Horgan,
  Dhruva TB, Alistair Muldal, Nicolas Heess, and Timothy Lillicrap.
\newblock Distributional policy gradients.
\newblock In {\em International Conference on Learning Representations}, 2018.

\bibitem{bellemare2016unifying}
Marc~G Bellemare, Sriram Srinivasan, Georg Ostrovski, Tom Schaul, David Saxton,
  and Remi Munos.
\newblock Unifying count-based exploration and intrinsic motivation.
\newblock {\em arXiv preprint arXiv:1606.01868}, 2016.

\bibitem{burda2018exploration}
Yuri Burda, Harrison Edwards, Amos Storkey, and Oleg Klimov.
\newblock Exploration by random network distillation.
\newblock In {\em International Conference on Learning Representations}, 2019.

\bibitem{chen2020robust}
Bryan Chen, Alexander Sax, Gene Lewis, Iro Armeni, Silvio Savarese, Amir Zamir,
  Jitendra Malik, and Lerrel Pinto.
\newblock Robust policies via mid-level visual representations: An experimental
  study in manipulation and navigation.
\newblock {\em arXiv preprint arXiv:2011.06698}, 2020.

\bibitem{de2018integrating}
Tim de~Bruin, Jens Kober, Karl Tuyls, and Robert Babu{\v{s}}ka.
\newblock Integrating state representation learning into deep reinforcement
  learning.
\newblock {\em IEEE Robotics and Automation Letters}, 3(3):1394--1401, 2018.

\bibitem{Ghosh2020From}
Partha Ghosh, Mehdi S.~M. Sajjadi, Antonio Vergari, Michael Black, and Bernhard
  Scholkopf.
\newblock From variational to deterministic autoencoders.
\newblock In {\em International Conference on Learning Representations}, 2020.

\bibitem{goodfellow2014generative}
Ian~J Goodfellow, Jean Pouget-Abadie, Mehdi Mirza, Bing Xu, David Warde-Farley,
  Sherjil Ozair, Aaron Courville, and Yoshua Bengio.
\newblock Generative adversarial nets.
\newblock In {\em Proceedings of the 27th International Conference on Neural
  Information Processing Systems-Volume 2}, pages 2672--2680, 2014.

\bibitem{haarnoja2018soft}
Tuomas Haarnoja, Aurick Zhou, Pieter Abbeel, and Sergey Levine.
\newblock Soft actor-critic: Off-policy maximum entropy deep reinforcement
  learning with a stochastic actor.
\newblock In {\em International Conference on Machine Learning}, pages
  1861--1870. PMLR, 2018.

\bibitem{hafner2019learning}
Danijar Hafner, Timothy Lillicrap, Ian Fischer, Ruben Villegas, David Ha,
  Honglak Lee, and James Davidson.
\newblock Learning latent dynamics for planning from pixels.
\newblock In {\em International Conference on Machine Learning}, pages
  2555--2565. PMLR, 2019.

\bibitem{pmlr-v119-henaff20a}
Olivier Henaff.
\newblock Data-efficient image recognition with contrastive predictive coding.
\newblock In {\em Proceedings of the 37th International Conference on Machine
  Learning}, Proceedings of Machine Learning Research. PMLR, 2020.

\bibitem{higgins2016beta}
Irina Higgins, Loic Matthey, Arka Pal, Christopher Burgess, Xavier Glorot,
  Matthew Botvinick, Shakir Mohamed, and Alexander Lerchner.
\newblock beta-vae: Learning basic visual concepts with a constrained
  variational framework.
\newblock In {\em International Conference on Learning Representations (ICLR)},
  2016.

\bibitem{higgins2017darla}
Irina Higgins, Arka Pal, Andrei Rusu, Loic Matthey, Christopher Burgess,
  Alexander Pritzel, Matthew Botvinick, Charles Blundell, and Alexander
  Lerchner.
\newblock Darla: Improving zero-shot transfer in reinforcement learning.
\newblock In {\em International Conference on Machine Learning}, pages
  1480--1490. PMLR, 2017.

\bibitem{houthooft2016vime}
Rein Houthooft, Xi~Chen, Yan Duan, John Schulman, Filip De~Turck, and Pieter
  Abbeel.
\newblock Vime: Variational information maximizing exploration.
\newblock {\em arXiv preprint arXiv:1605.09674}, 2016.

\bibitem{jonschkowski2015learning}
Rico Jonschkowski and Oliver Brock.
\newblock Learning state representations with robotic priors.
\newblock {\em Autonomous Robots}, 39(3):407--428, 2015.

\bibitem{kingma2014adam}
Diederik~P Kingma and Jimmy Ba.
\newblock Adam: A method for stochastic optimization.
\newblock {\em arXiv preprint arXiv:1412.6980}, 2014.

\bibitem{Kingma2014}
Diederik~P. Kingma and Max Welling.
\newblock {Auto-Encoding Variational Bayes}.
\newblock In {\em International Conference on Learning Representations (ICLR)},
  2014.

\bibitem{lange2010deep}
Sascha Lange and Martin Riedmiller.
\newblock Deep auto-encoder neural networks in reinforcement learning.
\newblock In {\em The 2010 International Joint Conference on Neural Networks
  (IJCNN)}, pages 1--8. IEEE, 2010.

\bibitem{laskin2020curl}
Michael Laskin, Aravind Srinivas, and Pieter Abbeel.
\newblock Curl: Contrastive unsupervised representations for reinforcement
  learning.
\newblock In {\em International Conference on Machine Learning}, pages
  5639--5650. PMLR, 2020.

\bibitem{lee2019making}
Michelle~A Lee, Yuke Zhu, Krishnan Srinivasan, Parth Shah, Silvio Savarese,
  Li~Fei-Fei, Animesh Garg, and Jeannette Bohg.
\newblock Making sense of vision and touch: Self-supervised learning of
  multimodal representations for contact-rich tasks.
\newblock In {\em 2019 International Conference on Robotics and Automation
  (ICRA)}, pages 8943--8950. IEEE, 2019.

\bibitem{lesort2018state}
Timoth{\'e}e Lesort, Natalia D{\'\i}az-Rodr{\'\i}guez, Jean-Franois Goudou, and
  David Filliat.
\newblock State representation learning for control: An overview.
\newblock {\em Neural Networks}, 108:379--392, 2018.

\bibitem{noroozi2016unsupervised}
Mehdi Noroozi and Paolo Favaro.
\newblock Unsupervised learning of visual representations by solving jigsaw
  puzzles.
\newblock In {\em European conference on computer vision}, pages 69--84.
  Springer, 2016.

\bibitem{oord2018representation}
Aaron van~den Oord, Yazhe Li, and Oriol Vinyals.
\newblock Representation learning with contrastive predictive coding.
\newblock {\em arXiv preprint arXiv:1807.03748}, 2018.

\bibitem{osband2016deep}
Ian Osband, Charles Blundell, Alexander Pritzel, and Benjamin Van~Roy.
\newblock Deep exploration via bootstrapped dqn.
\newblock {\em arXiv preprint arXiv:1602.04621}, 2016.

\bibitem{ostrovski2017count}
Georg Ostrovski, Marc~G Bellemare, A{\"a}ron Oord, and R{\'e}mi Munos.
\newblock Count-based exploration with neural density models.
\newblock In {\em International conference on machine learning}, pages
  2721--2730. PMLR, 2017.

\bibitem{pairet2019learning}
{\`E}ric Pairet, Paola Ard{\'o}n, Frank Broz, Michael Mistry, and Yvan
  Petillot.
\newblock Learning and generalisation of primitives skills towards robust
  dual-arm manipulation.
\newblock {\em arXiv preprint arXiv:1904.01568}, 2019.

\bibitem{NEURIPS2019_9015}
Adam Paszke, Sam Gross, Francisco Massa, Adam Lerer, James Bradbury, Gregory
  Chanan, Trevor Killeen, Zeming Lin, Natalia Gimelshein, Luca Antiga, Alban
  Desmaison, Andreas Kopf, Edward Yang, Zachary DeVito, Martin Raison, Alykhan
  Tejani, Sasank Chilamkurthy, Benoit Steiner, Lu~Fang, Junjie Bai, and Soumith
  Chintala.
\newblock Pytorch: An imperative style, high-performance deep learning library.
\newblock In {\em Advances in Neural Information Processing Systems}. 2019.

\bibitem{pathak2017curiosity}
Deepak Pathak, Pulkit Agrawal, Alexei~A Efros, and Trevor Darrell.
\newblock Curiosity-driven exploration by self-supervised prediction.
\newblock In {\em International Conference on Machine Learning}, pages
  2778--2787. PMLR, 2017.

\bibitem{schmidhuber1991curious}
Juergen Schmidhuber.
\newblock Curious model-building control systems.
\newblock In {\em Proc. international joint conference on neural networks},
  pages 1458--1463, 1991.

\bibitem{seo2021state}
Younggyo Seo, Lili Chen, Jinwoo Shin, Honglak Lee, Pieter Abbeel, and Kimin
  Lee.
\newblock State entropy maximization with random encoders for efficient
  exploration.
\newblock {\em arXiv preprint arXiv:2102.09430}, 2021.

\bibitem{stooke2020decoupling}
Adam Stooke, Kimin Lee, Pieter Abbeel, and Michael Laskin.
\newblock Decoupling representation learning from reinforcement learning.
\newblock {\em arXiv preprint arXiv:2009.08319}, 2020.

\bibitem{strubell2019energy}
Emma Strubell, Ananya Ganesh, and Andrew McCallum.
\newblock Energy and policy considerations for deep learning in nlp.
\newblock {\em arXiv preprint arXiv:1906.02243}, 2019.

\bibitem{sutton2018reinforcement}
Richard~S Sutton and Andrew~G Barto.
\newblock {\em Reinforcement learning: An introduction}.
\newblock MIT press, 2018.

\bibitem{tassa2018deepmind}
Yuval Tassa, Yotam Doron, Alistair Muldal, Tom Erez, Yazhe Li, Diego de~Las
  Casas, David Budden, Abbas Abdolmaleki, Josh Merel, Andrew Lefrancq, et~al.
\newblock Deepmind control suite.
\newblock {\em arXiv preprint arXiv:1801.00690}, 2018.

\bibitem{van2016stable}
Herke Van~Hoof, Nutan Chen, Maximilian Karl, Patrick van~der Smagt, and Jan
  Peters.
\newblock Stable reinforcement learning with autoencoders for tactile and
  visual data.
\newblock In {\em 2016 IEEE/RSJ international conference on intelligent robots
  and systems (IROS)}, pages 3928--3934. IEEE, 2016.

\bibitem{watter2015embed}
Manuel Watter, Jost~Tobias Springenberg, Joschka Boedecker, and Martin
  Riedmiller.
\newblock Embed to control: A locally linear latent dynamics model for control
  from raw images.
\newblock {\em arXiv preprint arXiv:1506.07365}, 2015.

\bibitem{wu2018unsupervised}
Zhirong Wu, Yuanjun Xiong, Stella~X Yu, and Dahua Lin.
\newblock Unsupervised feature learning via non-parametric instance
  discrimination.
\newblock In {\em Proceedings of the IEEE Conference on Computer Vision and
  Pattern Recognition}, pages 3733--3742, 2018.

\bibitem{yarats2019improving}
Denis Yarats, Amy Zhang, Ilya Kostrikov, Brandon Amos, Joelle Pineau, and Rob
  Fergus.
\newblock Improving sample efficiency in model-free reinforcement learning from
  images.
\newblock {\em arXiv preprint arXiv:1910.01741}, 2019.

\end{thebibliography}

%\appendix
\appendices
\section{Implementation Details \& Hyperparameters}
\label{app:hyper}

For the encoder and decoder, we employ the architecture from~\cite{yarats2019improving}. Both consist of four convolutional layers with $3 \times 3$ kernels and $32$ channels and use ReLU activations, except of the final deconvolution layer. Both networks use a stride of 1 for each layer except of the first of the encoder and the last of the decoder, which use stride 2.

For all shown experiments, we train with multiple seeds for each task. At the beginning of each seed, we pretrain the models with 1000 samples which we collect by rolling out random actions. Afterwards, we evaluate the model every 10 thousand environment steps over 10 episodes and report the average reward. The total number of episodes depends on the complexity of the task. All hyperparameters used in our experiments are summarized in Table~\ref{tab:hyper}.

\begin{table}[!htb]
\normalsize
\begin{center}
\begin{tabular}{ l|c } 
 \hline
 Parameter & Setting \\ 
 \hline
 Batch size & 128 \\ 
 Replay buffer capacity & 80000 \\ 
 Discount $\gamma$ & 0.99 \\
 Hidden dimension & 1024 \\
 Curious exploration probability $p_c$ & 0.2 \\
 Observation size & $84\times84\times3$ \\
 Frames stacked & 3 \\
 Critic learning rate & $10^{-3}$ \\
 Critic target update frequency & 2 \\
 Critic soft target update rate $\tau$ & 0.01 \\
 Actor learning rate & $10^{-3}$ \\
 Actor update frequency & 2 \\
 Actor log std bounds & [-10, 2] \\
 Autoencoder learning rate & $10^{-3}$ \\
 Decoder update frequency & 1 \\
 Temperature learning rate & $10^{-4}$ \\
 
 Init temperature & 0.1 \\
 
 \hline
\end{tabular}
\caption{The hyperparameters used in our experiments.}
\label{tab:hyper}
\end{center}
\end{table}

%\clearpage
\section{Additional Results}
\label{app:res}

Here we show the exact values obtained for the SRL error visitation experiment in section~\ref{sec:results}. These values were roughly illustrated in figure~\ref{fig:ae_error} and exactly shown in Table~\ref{table:ae_error}.

\begin{table}[htb!]
\normalsize
\begin{center}
%\resizebox{\textwidth}{!}{
\begin{tabular}{llccc}
\hline
\textbf{method/env} & \textbf{Vals} & \textbf{random} & \textbf{sac\_ae} & \textbf{cure} \\ \hline
\multirow{3}{*}{\textbf{reacher\_easy}} & \textbf{Min} & 0.0001 & 0.0001 & 0.0390 \\
 & \textbf{Mean} & 0.0002 & 0.0004 & 0.0399 \\
 & \textbf{Max} & 0.0003 & 0.0007 & 0.0424 \\ \hline
\multirow{3}{*}{\textbf{ball\_in\_cup}} & \textbf{Min} & 0.0005 & 0.0003 & 0.0762 \\
 & \textbf{Mean} & 0.0006 & 0.0005 & 0.0774 \\
 & \textbf{Max} & 0.0008 & 0.0007 & 0.0783 \\ \hline
\multirow{3}{*}{\textbf{cartpole\_swingup}} & \textbf{Min} & 0.0002 & 0.0002 & 0.0531 \\
 & \textbf{Mean} & 0.0002 & 0.0002 & 0.0540 \\
 & \textbf{Max} & 0.0003 & 0.0003 & 0.0548 \\ \hline
\multirow{3}{*}{\textbf{finger\_spin}} & \textbf{Min} & 0.0002 & 0.0004 & 0.0755 \\
 & \textbf{Mean} & 0.0003 & 0.0012 & 0.0766 \\
 & \textbf{Max} & 0.0004 & 0.0015 & 0.0774 \\ \hline
\multirow{3}{*}{\textbf{finger\_turn\_easy}} & \textbf{Min} & 0.0003 & 0.0003 & 0.0749 \\
 & \textbf{Mean} & 0.0004 & 0.0005 & 0.0759 \\
 & \textbf{Max} & 0.0006 & 0.0008 & 0.0769 \\
 \hline
\multirow{3}{*}{\textbf{reacher\_hard}} & \textbf{Min} & 0.0002 & 0.0002 & 0.0393 \\
 & \textbf{Mean} & 0.0003 & 0.0004 & 0.0397 \\
 & \textbf{Max} & 0.0008 & 0.0006 & 0.0404
 \\
 \hline
 \\
\end{tabular}%}
\caption{Mean, minimum and maximum SRL error encountered per step when using three different agents on six deepmind control suite tasks.}
\label{table:ae_error}
\end{center}
\end{table}

\end{document}